\pdfoutput=1

\documentclass[twocolumn]{article}
\usepackage[margin=1in]{geometry}
\usepackage{natbib}
\usepackage{authblk}

\usepackage{times}
\usepackage{latexsym}

\usepackage[T1]{fontenc}

\usepackage[utf8]{inputenc}

\usepackage{microtype}

\usepackage{inconsolata}

\usepackage{booktabs}
\usepackage{graphicx}

\usepackage{hyperref}

\title{Auto-Patching: Enhancing Multi-Hop Reasoning in Language Models}

\author[1]{Aviv Jan}
\author[1]{Dean Tahory}
\author[1]{Omer Talmi}
\author[1]{Omar Abo Mokh}

\affil[1]{Tel Aviv University\\
\texttt{\{avivjan, deantahory, omertalmi, omarabomokh\}@mail.tau.ac.il}}

\date{Jan 31, 2025}
  
\begin{document}
\maketitle

\begin{center}
\section*{Abstract}
\end{center}
\vspace{0.5em}

    Multi-hop questions still stump large language models (LLMs), which struggle to link information across multiple reasoning steps. We introduce Auto-Patch, a novel method that dynamically patches hidden states during inference to enhance multi-hop reasoning in LLMs. Building on the PatchScopes framework, Auto-Patch selectively modifies internal representations using a learned classifier. Evaluated on the MuSiQue dataset, Auto-Patch improves the solve rate from 18.45\% (baseline) to 23.63~$\pm$~0.7\% (3 runs), narrowing the gap to Chain-of-Thought prompting (27.44\%). Our results highlight the potential of dynamic hidden state interventions for advancing complex reasoning in LLMs.

\vspace{1em}
\noindent \textbf{Keywords:} Large Language Models, Multi-Hop Reasoning, Model Interpretability, PatchScopes, Dynamic Patching

\footnotetext[1]{Code available at: \href{https://github.com/omertalmi5/Auto-Patching}{github.com/omertalmi5/Auto-Patching}}

\section{Introduction}

\begin{figure}[!ht]
  \centering
  \includegraphics[width=0.5\textwidth]{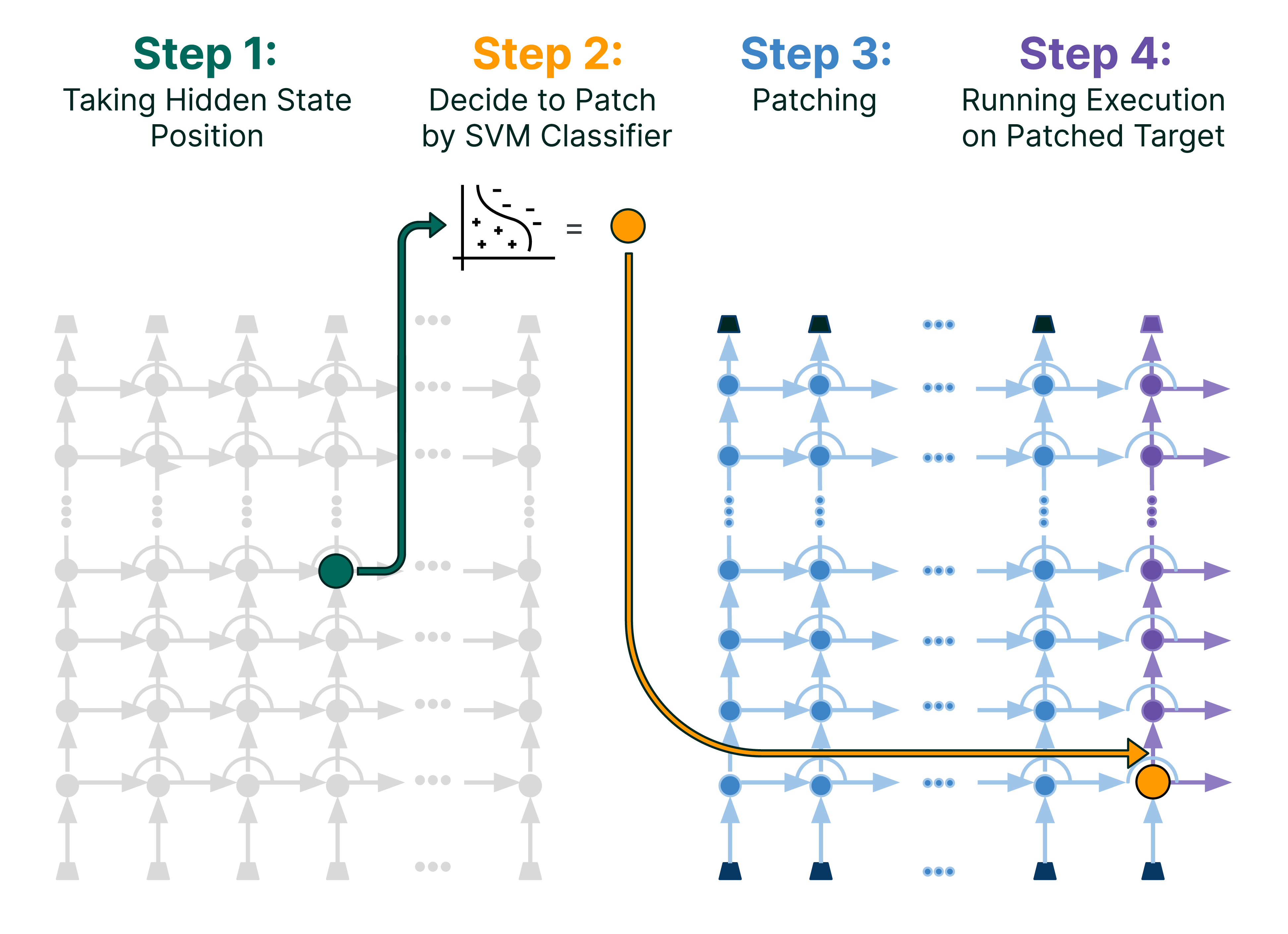}
  \caption{Illustration of the Auto Patch framework used for enhancing multi-hop reasoning in language models. The process involves the following steps:
    \textbf{Step~1}: Select the hidden state from the source layer that will be considered for patching.
    \textbf{Step~2}: The SVM classifier evaluates whether the selected hidden state should be patched.
    \textbf{Step~3}: If patching is needed, the hidden state is transferred to the same position in the target layer.
    \textbf{Step~4}: Execute the forward pass on the patched target, incorporating the patched hidden state into the final output generation. This dynamic adjustment improves the model's ability to reason across multiple steps.}
  \label{fig:autopatch}
\end{figure}

The field of natural language processing (NLP) has been profoundly transformed by the advent of large language models (LLMs) such as BERT \cite{devlin2019bert}, GPT-3 \cite{brown2020language}, and LLaMA \cite{touvron2023llama}. These models have set new benchmarks for a variety of tasks, significantly improving our ability to understand and generate human language. Despite these advancements, LLMs continue to face challenges in handling complex reasoning tasks, especially multi-hop questions, which require synthesizing information from multiple sources to arrive at a correct answer.

Multi-hop reasoning involves answering questions that necessitate linking disparate pieces of information across different segments. For instance, the question \textbf{``Who is the current CEO of the company that created Alexa?''} has two parts. Although a language model can easily know which company created Alexa (Amazon) and who the CEO of Amazon is, it can show difficulties in answering the entire complex question. This might require first identifying a piece of data and then using that information to find the final answer. Traditional approaches to improving language models' performance on such tasks include:

\begin{enumerate}
    \item \textbf{Chain-of-Thought (CoT) Prompting}: This method guides the model to generate explicit intermediate reasoning steps. CoT prompting helps break down complex problems into manageable steps, significantly enhancing performance on tasks such as arithmetic and common sense reasoning. However, CoT often requires extensive fine-tuning and precise prompt engineering, which can be computationally expensive and may not generalize well without considerable manual effort \cite{wei2023chainofthought}, \cite{zhang2022automatic}.
    
    \item \textbf{Memory-Augmented Networks}: These networks integrate external memory to store and retrieve information across multiple hops. Although they are effective in handling complex reasoning using stored data, managing this memory efficiently can be challenging. The added computational overhead and complexity can limit their scalability and overall performance \cite{wang-etal-2023-towards}.
    
    \item \textbf{Graph-Based Approaches}: These methods use graph neural networks to represent and reason over relationships in the question and relevant information. Graph-based approaches excel in capturing relationships between entities and steps in multi-hop questions. However, they are often difficult to scale and integrate seamlessly with LLMs, requiring specialized infrastructure and significant computational resources \cite{de-cao-etal-2019-question}.
\end{enumerate}

Despite their success, these methods come with limitations, such as increased computational demands, complexity in implementation, and sometimes only modest performance gains. This study explores how the reasoning capabilities of LLMs can be enhanced through a novel approach inspired by PatchScopes \cite{ghandeharioun2024patchscopes}. PatchScopes is a modular framework that allows one to inspect and manipulate hidden states in language models. By "patching" or adjusting these hidden states during the model's computation, PatchScopes can refine internal representations and improve reasoning capabilities.

\subsection*{Objective}
In this paper, our objective is to improve the performance of the LLaMA 2 model (7B version) on 2-hop question answering tasks. Leveraging the PatchScopes framework, we introduce the Auto Patch method, which dynamically adjusts hidden states during the model's inference process. Our goal is to improve the model’s ability to synthesize information across multiple reasoning steps.

\subsection*{Contribution}
We present a novel approach that utilizes a classifier to determine which hidden states should be patched to enhance the model’s performance. By selectively patching these states, we achieved significant improvements over the baseline in handling 2-hop questions. This research underscores the practical application of dynamic hidden state manipulation via PatchScopes, providing insights into optimizing LLM performance and advancing model interpretability.
\section{Related Work}
Mechanistic interpretability (MI) aims to unravel complex machine learning models by reverse engineering their internal mechanisms down to human-understandable algorithms (Geiger et al., 2021; Olah, 2022; Wang et al., 2023). With such understanding, we can better identify and fix model errors (Vig et al., 2020), steer model outputs (Li et al., 2023b) and explain emergent behaviors (Nanda et al., 2023a; Barak et al., 2022). 
Intervening during inference can, in some cases, increase the performance of the model.(link in docs).

\subsection*{Multi Hop}
Multi-hop reasoning involves answering questions that require understanding and linking information from two or more separate data points. For example, a question like "Who is the president of the company that was founded by Elon Musk?" requires the system first to recognize the company Elon Musk founded and then determine the current president of that company. While a language model may be capable of correctly answering each step independently, it could still fail at processing the connection between different steps, resulting in an incorrect prediction.
A notable precursor in addressing complex multi-step reasoning tasks is the Chain of Thought (CoT) approach. CoT prompts language models to generate intermediate reasoning steps explicitly before arriving at a final answer, mimicking human problem-solving processes. This method has demonstrated enhanced performance on reasoning tasks by making the reasoning process explicit, which helps in aligning the model's processing steps with the logical flow required to answer multi-hop questions.

\subsection*{Patchscopes}
Patchscopes, introduced by Gandeharioun et al., presents a modular framework for interpreting the hidden representations of large language models (LLMs).
This framework capitalizes on the advanced text generation capabilities of LLMs to explicate their internal representations in a human-understandable format. By dynamically "patching" transformed hidden states into different stages of the model's inference process, Patchscopes facilitates a detailed inspection of model behaviors across various layers. It effectively unifies and extends previous interpretability methods, which typically involve projecting representations to the vocabulary space or directly intervening in the model's computation process. A key innovation of Patchscopes is its ability to utilize more capable models to elucidate the operations of less complex ones, thereby enhancing the expressiveness and robustness of the interpretations. This approach not only addresses limitations such as the inability to probe early layers but also introduces novel capabilities like correcting multi-hop reasoning errors.
\newline \newline
However, Patchscopes employs a method that necessitates manual separation of prompts into single components for multi-hop queries in order to find the vectors to patch - an approach that may not be practical in real-world applications. In this work we advance the Patchscopes methodology by developing an automated patching process that eliminates the need for manual prompt separation, thereby bridging a significant gap in the practical deployment of model interpretability techniques.

\section{Methodology}
This section details the improved methods used to enhance the multi-hop reasoning abilities of the LLaMA 2 model by integrating an automated patching process within the PatchScopes framework. Our approach removes the need for manually separating prompts into components, making it more practical for real-world use.

\subsection{Advancing PatchScopes with Auto Patch}
PatchScopes is a key framework for exploring and adjusting hidden states within large language models (LLMs), helping to refine model behaviors and improve reasoning capabilities. Although promising, traditional PatchScopes often require extensive manual work, which limits their use outside of research settings. Our development extends this framework through the Auto Patch method, which uses classifiers to automatically decide which hidden states to patch. This not only makes the patching process easier but also makes the model more scalable and efficient, especially in complex multi-hop reasoning tasks. The main idea behind our method is that the decision to apply a patch can be effectively learned and optimized, thus improving the model’s accuracy and functionality.

\subsection{Data Preparation for Classifier Training}
The data preparation process for training the classifier involves the following steps:
For each experiment, we selected specific layers "source layer" (from the first run) and "target layer" (from the second run) within the language model's hidden states. Each prompt in the dataset was processed position by position through the language model. For each position i in the "target layer", we patched the hidden state to the corresponding position i in the "source layer".
After applying the patch on each position, we evaluated the model's output to determine if patching improved the likelihood of generating the correct answer. If patching resulted in a higher probability of the correct answer being generated, we labeled the hidden state values at position i as True (indicating they should be used for patching). If there was no improvement or a decrease in performance, we labeled the hidden state values as False.
The final dataset for classifier training consisted of the hidden state values as features, with each data point labeled as either True or False based on the effectiveness of the patching process. We trained a separate classifier for each pair of layers "source layer" and "target layer" to predict whether patching specific positions from "target layer" to "source layer" would enhance the two-hop question-answering performance.

\subsection{Classifier Training for Auto Patching}
To support the automatic patching process, we trained a Support Vector Machine (SVM) classifier using a specific kernel to improve decision-making. Training involved extracting features from interactions between various layers, where hidden states from higher layers were experimentally replaced with those from lower layers to measure their impact on performance. This method allowed us to train the SVM classifier effectively, enabling it to dynamically determine the best positions for patching during model inference, thus removing the need for manual layer specification and focusing on enhancing reasoning through learned interventions.

\subsection{Auto Patching Framework Implementation}
At the heart of our methodology, the Auto Patching Framework dramatically changes how hidden states are manipulated during model inference. Initially, the model processes the prompt through a standard forward pass to capture the baseline hidden states. These states are then evaluated by the trained classifier to decide if and where patches are needed. If patching is beneficial, the model undergoes a second forward pass where selected hidden states are adjusted as recommended by the classifier. This dual-phase operation not only maintains the necessary context for accurate multi-hop reasoning but also adds a dynamic, learnable element to hidden state manipulation, greatly improving the model's performance on various prompts.

To enable this process, we developed specific patching code that started with manual patching to gain initial insights into the strategic manipulation of hidden states. This phase was crucial for determining the impactful interactions between layers and setting the groundwork for automated interventions.

Following these initial insights, we progressed to automating the patching process. The decision-making capabilities of our SVM classifier were seamlessly integrated into the model’s inference engine, enabling dynamic and automatic patching during the question-answering process. This integration ensures that the model can adjust hidden states in real-time, based on classifier predictions, optimizing performance without the need for manual intervention.

This comprehensive development and integration of patching code into the Auto Patching Framework not only streamlines the entire process but also enhances the model’s ability to handle complex reasoning tasks more efficiently.

\section{Experiments}

This section presents the experiments conducted to evaluate the effectiveness of our Auto Patch method. We provide a detailed description of the experimental setup, evaluation metrics, dataset characteristics, results, and comparative analysis with baseline and Chain-of-Thought (CoT) methods.

\subsection{Evaluation Metric}

To evaluate the performance of our method, we utilized solve rate metric:
The primary metric was the solve rate, defined as the proportion of correctly answered 2-hop questions out of the total questions.

\subsection{Dataset Description}
Our dataset contains 24,912 samples created from 1,024 prompts from MuSiQue \cite{deldjoo2021musique}. The structure of the prompts of MuSiQue is that the first hop uses the second hop, for example: "What award was received by the person who authored Missing Person?". Each sample representing a different execution of the Patchcopes method with patching from and to specific positions. For each prompt, we executed from source layer 15 to target layer 8, for each position in the prompt. The main features of each sample are: 'prompt\_source' and 'prompt\_target', the same and consist of a full two-hop question; 'position\_source' and 'position\_target', the same position that from and to; 'hop3', the correct answer to the two-hop question; 'generations\_patched', the LLM's response to the two-hop question; 'is\_correct\_patched', a boolean value indicating if the LLM generated the correct answer; and 'hidden\_rep', the hidden state patched during execution, represented as a float array of length 4096. 'hidden\_rep' is the input for the classifier, and 'is\_correct\_patched' is the label.
23\% of the samples have 'is\_correct\_patched' as true. This does not mean that the solve rate is 23\%, because long prompts occur in more samples.

\subsection{The Experiment Flow}
We receive 1,024 two-hop questions without any separation into hops. We pass them through the first inference in the LLM. Then, each hidden state from layer 15 passes through the classifier. If the classifier returns true, in the second inference, we replace it with the hidden state in the same position in layer 8. We then complete the second inference and check if it returns the correct answer.

\subsection{Results and Analysis}

The results of our experiments demonstrate the impact of the Auto Patch method on model performance compared to the baseline and Chain-of-Thought (CoT) methods. Table~\ref{tab:results} summarizes the solve rate of each approach on the 2-hop questions from the MuSiQue dataset.

\begin{table}[h!]
    \centering
    \resizebox{\columnwidth}{!}{
        \begin{tabular}{|c|c|}
            \hline
            \textbf{Method} & \textbf{Solve Rate (\%)} \\
            \hline
            Baseline (LLaMA 2) & 18.45 \\
            \hline
            Chain-of-Thought (CoT) & 27.44 \\
            \hline
            Auto Patch (Ours) & 23.63 \\
            \hline
        \end{tabular}
    }
    \caption{Comparison of solve rate on MuSiQue 2-hop questions.}
    \label{tab:results}
\end{table}

\textbf{Auto Patch (Ours):} Our Auto Patch method achieved a solve rate of 23.63~$\pm$~0.7\%, significantly outperforming the baseline but not surpassing the CoT method.

To determine optimal positions for patching, we employed an SVM classifier from the sklearn library, utilizing a Radial Basis Function (RBF) kernel. Given the imbalanced nature of the dataset, we applied the SMOTETomek method for balancing, which involved both sampling and generating the minority class. Standardization was also part of the preprocessing to enhance the classifier's performance.

The classifier achieved an accuracy of 0.81. Table~\ref{tab:svm_report} provides a detailed classification report for the SVM classifier's performance at the source layer (layer 15).

\begin{table}[h!]
    \centering
    \resizebox{\columnwidth}{!}{
        \begin{tabular}{lcccc}
            \toprule
            \textbf{} & \textbf{Precision} & \textbf{Recall} & \textbf{F1-Score} & \textbf{Support} \\
            \midrule
            False & 0.84 & 0.92 & 0.88 & 3808 \\
            True & 0.64 & 0.45 & 0.53 & 1175 \\
            \midrule
            \textbf{Accuracy} & \multicolumn{4}{c}{0.81} \\
            \textbf{Macro Avg} & 0.74 & 0.69 & 0.70 & 4983 \\
            \textbf{Weighted Avg} & 0.80 & 0.81 & 0.80 & 4983 \\
            \bottomrule
        \end{tabular}
    }
    \caption{Classification Report for the classifier}
    \label{tab:svm_report}
\end{table}

As shown in Table~\ref{tab:svm_report}, The accuracy of the classifier was 0.81. While the overall precision and recall on the data after the preprocessing suggest acceptable performance. The classifier performed with high precision and recall for the 'False' class compared to the 'True' class, the majority class ('False') is predicted more accurately. On the real data, the model predicts \textit{True} for almost all positions. All samples have one or two positions that are not patched. Most of positions that not patched, their tokens are: start of sentence symbol <s> (can come with the real prompt after it and many <unk> before it), or unknown symbol <unk> or single dot "." , as shown in Figure 2. The classifier learns the structure of the sentence and mostly does not patch the first token. Intuitively, we can view this as patching only the first token from layer 8 to layer 13. The intuition for patching most of the positions is that the second hop should influence the first hop in our dataset.

\begin{figure}[!ht]
    \centering
    \includegraphics[width=\columnwidth]{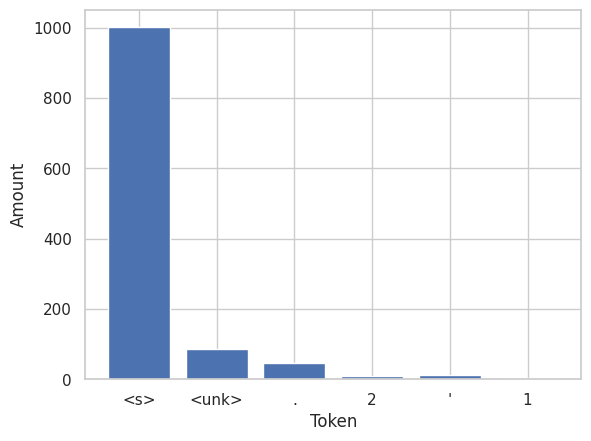}
    \caption{Histogram showing the tokens of not patched positions.}
    \vspace{-.5cm}
\end{figure}
\subsection{Additional Experiments}
To measure the effect of choosing layers "source layer" and "target layer" on the results, we conducted two additional experiments on smaller data.
These experiments were performed on 3,114 samples created from 128 prompts, following all the steps outlined in the main experiment.

\subsubsection*{Experiment 1: Effect of the Source Layer}
In this experiment, we investigated the impact of the "source layer" selection on the results. To maintain consistency, we fixed the distance between the "source layer" and the "target layer" at 5 layers. As we varied the "source layer," the "target layer" was adjusted correspondingly. For instance, if the "source layer" was set to 10, the "target layer" was positioned at 5; similarly, if the "source layer" was 11, the "target layer" was at 6, and so forth. This allowed us to evaluate how different "source layer" choices influenced the performance of the patching process.

\subsubsection*{Experiment 2: Effect of the Distance Between Layers}
This experiment focused on understanding how varying the distance between the "source layer" and "target layer" affects the results. We began with the "source layer" set at 16 and the "target layer" at 14. Gradually, we increased the distance by incrementing the "source layer" by 1 and decrementing the "target layer" by 1 in each step. For example, we used "source layer" = 16 and "target layer" = 14, then adjusted to "source layer" = 17 and "target layer" = 13, continuing this pattern. This approach helped us assess how different distances between layers influence the effectiveness of the patching process.

\subsubsection*{Results}
\begin{figure}[!ht]
    \centering
    \includegraphics[width=\columnwidth]{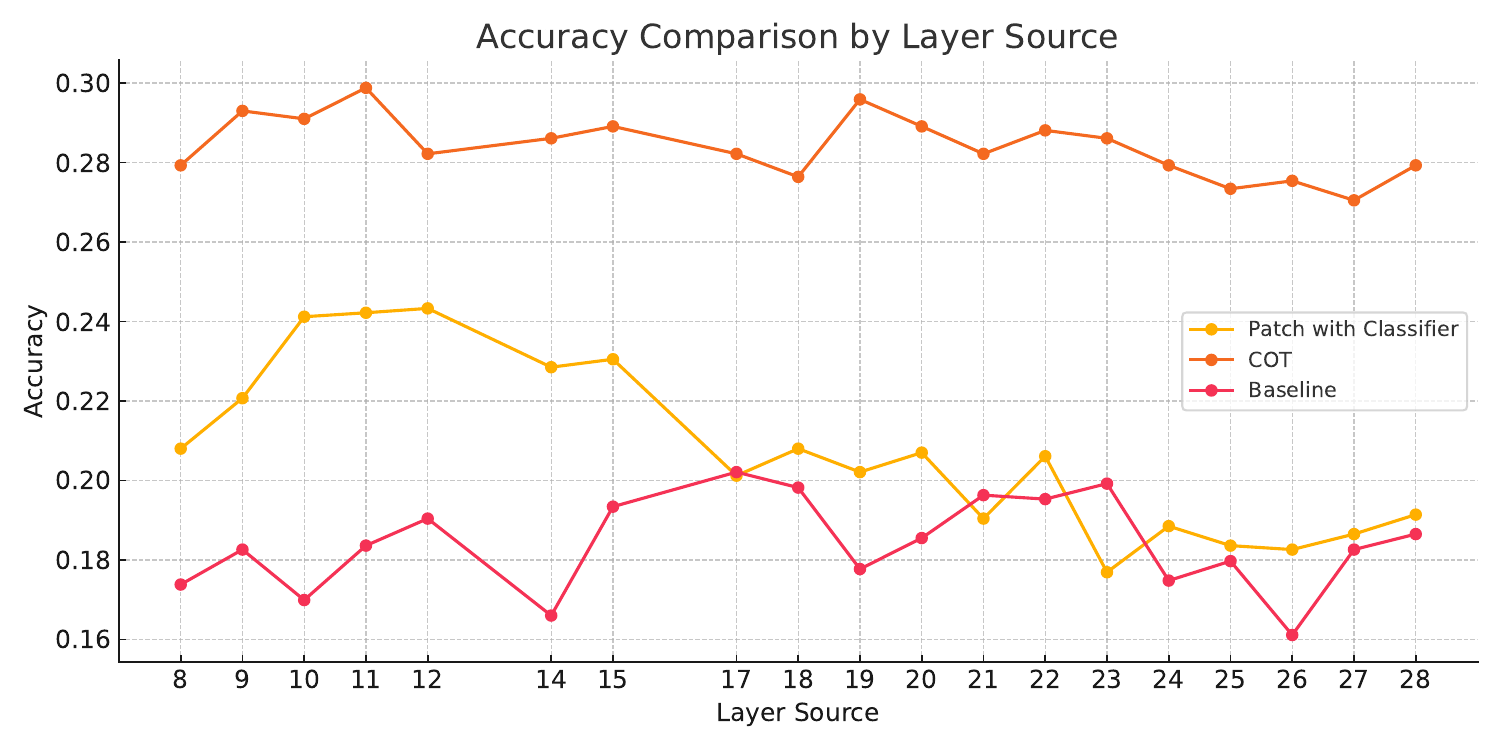}
    \caption{The x-axis represents the source layer, and the y-axis represents the accuracy of the patching process while keeping the distance between source layer and target layer constant at 5 layers.}
    \vspace{-.5cm}
\end{figure}
\begin{figure}[!ht]
    \centering
    \includegraphics[width=\columnwidth]{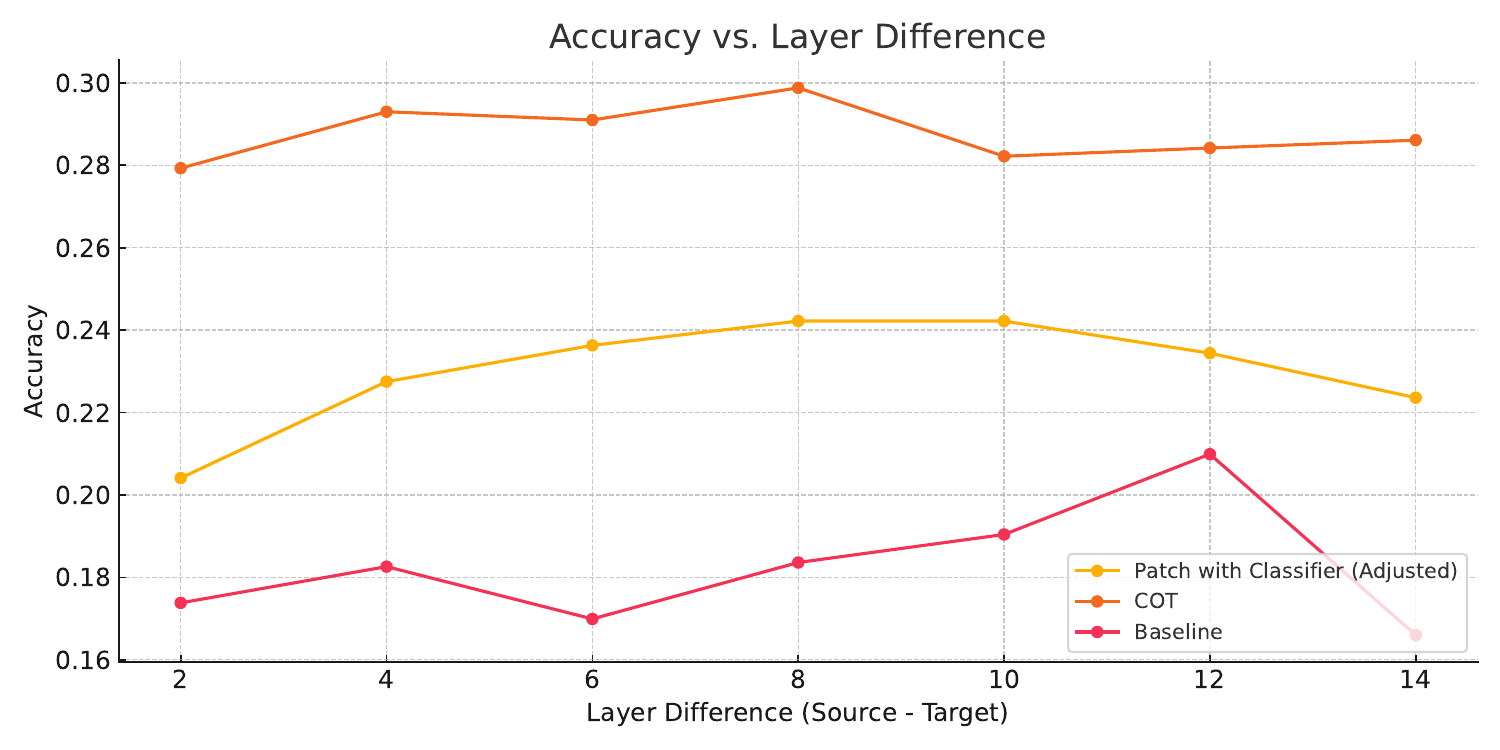}
    \caption{The x-axis represents the increasing distance between layers source and target, and the y-axis represents the accuracy of the patching process.}
    \vspace{-.5cm}
\end{figure}
\paragraph{Observation}
In the first graph, we see that the accuracy was low initially, peaked around layers 10-12, and then started to decline.
\paragraph{Explanation} The initial layers of a language model are involved in contextualizing the input tokens, forming intermediate representations that are still processing and integrating the input information. Around layers 10-12, the hidden states carry a balanced amount of processed information that is both specific enough to be useful for prediction tasks and general enough to benefit from patching. This makes them optimal for patching, resulting in the highest accuracy. As we move beyond these layers, the representations become over-specialized for specific next-token predictions, reducing their flexibility and effectiveness for holding context data from the first hop. This trend aligns with the findings in the Patchscopes research \cite{ghandeharioun2024patchscopes}
\paragraph{for second experiment} 
As the distance between layers source and target increases, the patching process initially benefits the two-hop question answering performance. This is because greater separation allows the patched layers to provide more comprehensive and processed contextual information, which is crucial for multi-hop reasoning tasks. The intermediate layers between source and target help bridge the information, enhancing the overall understanding and accuracy of the two-hop questions. However, as the distance becomes too large, the alignment and consistency between the layers may deteriorate, leading to a slight decline in performance

\subsection{Discussion}
The effectiveness of the Auto Patch method can be attributed to several key factors that enhance the multi-hop reasoning capabilities of large language models (LLMs).

\paragraph{Intuition Behind Patching} The Auto Patch method uses dynamic hidden state patching to reintroduce and modify information at specific layers during inference. This process refines the model’s understanding, maintaining coherence and context across reasoning steps of the different hops. It allows the model to link separate pieces of information, which is essential for multi-hop reasoning tasks.

\paragraph{Increased Computational Utilization} Similar to Chain-of-Thought (CoT) prompting, which utilizes more compute by guiding the model through intermediate reasoning steps, the Auto Patch method enhances computational engagement. By dynamically adjusting hidden states, the method directs computational resources towards synthesizing complex, multi-step information, improving accuracy and depth of reasoning.

\paragraph{Classifier's Role in Effective Patching} The integration of a Support Vector Machine (SVM) classifier is crucial in determining when and where patches should be applied. The classifier identifies scenarios where patching hidden states is most beneficial, allowing dynamic adaptation and improvement of internal representations. This showcases a learned strategy for enhancing multi-hop reasoning.

\paragraph{Comparison with Random Classification} The classifier’s effectiveness is highlighted by its performance compared to a random classification baseline. The random classification serves as a control, demonstrating that the observed improvements are due to the targeted, learned patching strategy rather than arbitrary interventions. This comparison underscores the classifier's role as a learned, intelligent component in the Auto Patch framework.

\subsection{Unsuccessful Experiments}
In the course of developing and evaluating the Auto Patch method, several experiments did not yield the anticipated improvements. These insights are crucial for guiding future research and refining our approach.

\paragraph{Positional Patch Adjustments} One explored avenue was concatenating the position number to the hidden state that was fed to the classifier. The hypothesis was that the position number might add information to the classifier about the structure of the sentence. However, these experiments showed the same results. The suggestion is that the hidden states in LLaMA2 already incorporate positional information through the initial addition of position embeddings to the token embeddings. This integration means that the positional context is inherently preserved in the hidden states, rendering the explicit addition of position numbers unnecessary for the classifier.

\section{Conclusion}
In this paper, we introduced the Auto Patch method, designed to enhance multi-hop reasoning in large language models by dynamically adjusting hidden states. Our results on the MuSiQue dataset show improved performance on 2-hop questions.

Future work could explore context-aware classifiers that incorporate neighboring states for better decision-making. Additionally, applying patches to the same positions across layers may not be optimal; developing adaptive strategies for layer and position selection could enhance effectiveness. Finally, expanding testing to a broader range of datasets and question types will be crucial to evaluate the method’s broader applicability.

Auto-Patch demonstrates the effectiveness of dynamic hidden state manipulation in advancing LLM reasoning. Our results suggest promising directions for future work in model interpretability and complex reasoning.

\section*{Acknowledgements}

We would like to express our deepest gratitude to Dr. Mor Geva, Maor Ivgi, and Daniela Gottesman for their invaluable guidance and support throughout this project. Their insights and expertise have significantly contributed to the successful completion of our research.

\section*{Limitations}
While the Auto Patch method improves multi-hop reasoning, it has notable limitations. First, the SVM classifier's decisions are based on isolated hidden states, which may ignore broader context. Second, the method patches identical positions across layers, limiting flexibility and potential task-specific optimization.

\section*{Future Work}

To further enhance the capabilities of our Auto Patch method in improving multi-hop reasoning in large language models, we propose the following avenues for future research:

\begin{enumerate}
    \item \textbf{Context-Aware Classifier Development:} The effectiveness of our current classifier, which receives input from a single hidden state at a given layer and position, is somewhat limited by its narrow view of the model's internal state. There is a clear indication that the decision to patch or not is significantly influenced by the actions (i.e., patching decisions) of neighboring states. To address this, we propose developing a new classifier model that can view and analyze a broader range of hidden states simultaneously. By incorporating more contextual information, such as the states of neighboring positions and even other related layers, the classifier would make more informed decisions that consider the inter-dependencies of patching actions. We believe that such a context-aware approach will significantly refine the precision of our patching interventions and    lead to notable improvements in model performance.
    
    \item \textbf{Advanced Classifier and Learning Algorithms:} Current implementations utilize a classifier that makes patching decisions based on isolated hidden states at specific layers and positions. Recognizing the potential for more nuanced decision-making, we plan to explore more sophisticated machine learning algorithms. Future iterations could involve deep learning models that can capture complex dependencies in data, potentially increasing the accuracy and efficiency of the patching decisions.

    \item \textbf{Optimization of Layer and Position Selection:} Our initial studies have highlighted the importance of selecting the appropriate layers and positions for effective patching. We intend to delve deeper into this aspect by investigating various strategies for selecting these parameters. This could involve developing heuristic algorithms or learning-based approaches that dynamically determine the most impactful layers and positions for patching, thereby maximizing the performance gains.

    \item \textbf{Expansion to Diverse Datasets and Question Types:} While our current experiments have been conducted on the MuSiQue dataset focusing on 2-hop questions, the versatility and robustness of the Auto Patch method can be further validated by extending these experiments to include a broader range of datasets and question types. This expansion will help ascertain the generalizability of our method across different contexts and benchmarks in natural language processing.

\end{enumerate}


\bibliographystyle{acl_natbib}

\end{document}